\documentclass[final,5p,times]{elsarticle}

\usepackage{graphicx}
\usepackage{amsmath}
\usepackage{amssymb}
\usepackage{hyperref}
\usepackage{lineno}
\usepackage{subcaption}
\usepackage{multirow}
\usepackage{dsfont}
\usepackage{float}  % Add this in your preamble
\usepackage[superscript]{cite}

\usepackage{booktabs} % for nice tables
\modulolinenumbers[5]
\newcommand{\suppMethodsSec}[1]{Supplementary Section #1}
\newcommand{\suppMethodsFig}[1]{Supplementary Figure #1}

\newcommand{\appSubsecHartford}{\suppMethodsSec{1.1}} 
\newcommand{\appSubsecStVinc}{\suppMethodsSec{1.2}} % Data Description and Curation
\newcommand{\appSeProgMatching}{\suppMethodsSec{2}} % Patient Identification and Exclusion
\newcommand{\appSecSampleWeight}{\suppMethodsSec{3}} % Target variables
\newcommand{\appFigDistPostMatching}{\suppMethodsFig{1}}

\newcommand{\appFigLovePlotRest}{\suppMethodsFig{3}}

\biboptions{sort&compress,super}

\journal{The Lancet Digital Health}

\begin{document}

\begin{frontmatter}

\title{An Interpretable AI Tool for SAVR vs TAVR in Low to Intermediate Risk Patients with Severe Aortic Stenosis}

\author{
Vasiliki Stoumpou, BSc$^{1}$,
Maciej Tysarowski, MD$^{2}$,
Talhat Azemi, MD$^{3,4}$,
Jawad Haider, MD$^{3,4}$,
Howard L. Haronian, MD$^{5}$,
Robert C. Hagberg, MD$^{3,4}$,
and Dimitris Bertsimas, PhD$^{1,6,*}$\\[1em]
\small $^{1}$Operations Research Center, Massachusetts Institute of Technology, Cambridge, MA, USA\\
\small $^{2}$Section of Cardiovascular Medicine, Yale School of Medicine, New Haven, CT, USA\\
\small $^{3}$Heart \& Vascular Institute, Hartford HealthCare, Hartford, CT, USA\\
\small $^{4}$Hartford HealthCare Research Institute, Hartford HealthCare, Hartford, CT, USA\\
\small $^{5}$Novant Health Heart \& Vascular Institute, Charlotte, NC, USA\\
\small $^{6}$Sloan School of Management, Massachusetts Institute of Technology, Cambridge, MA, USA\\[0.5em]
\small *Corresponding author: Dimitris Bertsimas \\ Sloan School of Management, Massachusetts Institute of Technology, Cambridge MA, 02142 \\ \texttt{dbertsim@mit.edu}
}

\begin{abstract}

\noindent 
Background.
Treatment selection for low to intermediate risk patients with severe aortic stenosis between surgical (SAVR) and transcatheter (TAVR) aortic valve replacement remains variable in clinical practice, driven by patient heterogeneity and institutional preferences. While existing models predict postprocedural risk, there is a lack of interpretable, individualized treatment recommendations that directly optimize long-term outcomes.

\noindent 
Methods.
We introduce an interpretable prescriptive framework that integrates prognostic matching, counterfactual outcome modeling, and an Optimal Policy Tree (OPT) to recommend the treatment minimizing expected 5-year mortality. Using data from Hartford Hospital (training, 236 patients) and St.~Vincent’s Hospital (external validation, 305 patients), we emulate randomization via prognostic matching and sample weighting and estimate counterfactual mortality under both SAVR and TAVR. The policy model, trained on these counterfactual predictions, partitions patients into clinically coherent subgroups and prescribes the treatment associated with lower estimated risk.

\noindent 
Findings.
The Optimal Policy Tree achieved high sensitivity (0·77) and moderate specificity (0·52) in the Hartford cohort, with 50\% concordance to real-world decisions. Counterfactual evaluation showed an estimated reduction in 5-year mortality of 20·3\% in Hartford and 13·8\% in St.~Vincent’s relative to real-life prescriptions, showing promising generalizability to unseen data from a different institution. The learned decision boundaries aligned with real-world outcomes and clinical observations: TAVR was favored in frailer or low-LVEF patients with larger sinus of Valsalva diameter, while SAVR was preferred in less frail patients with preserved LVEF or patients with small aortic annulus.

\noindent 
Interpretation.
Our interpretable prescriptive framework is, to the best of our knowledge, the first to provide transparent, data-driven recommendations for TAVR versus SAVR that improve estimated long-term outcomes both in an internal and external cohort, while remaining clinically grounded. The approach illustrates how the integration of methodologies to debias a cohort with policy tree learning can leverage observational data to generate actionable, individualized treatment policies, contributing toward a more systematic and evidence-based approach to precision medicine in structural heart disease.

\noindent 
Funding.
No dedicated external funding was received for this study. Computational resources and data infrastructure were provided by Hartford HealthCare as part of the research collaboration.

\end{abstract}


\end{frontmatter}

% -----------------------------
\section{Introduction}

\noindent
Aortic stenosis is one of the most common valvular heart diseases among older adults and is characterized by progressive narrowing of the aortic valve, leading to obstruction of blood flow from left ventricle to the aorta \citep{nkomo2006burden}. The condition often remains asymptomatic for many years but eventually manifests through symptoms such as chest pain, heart failure, or syncope. Once symptomatic, the disease becomes life-threatening if left untreated. Epidemiological studies estimate that aortic stenosis affects approximately 12·4\% of the elderly population, with severe forms present in about 3·4\% of older adults \citep{osnabrugge2013aortic}.

The definitive treatment for aortic stenosis is aortic valve replacement, performed either surgically (SAVR, Surgical Aortic Valve Replacement) or percutaneously (TAVR, Transcatheter Aortic Valve Replacement). Over the past decade, advances in TAVR technology have expanded its indications to progressively lower-risk and younger patients, challenging traditional surgical guidelines \citep{mack2023transcatheter, reyes2017transcatheter, sudhakaran2024treating}. Despite extensive studies showing promising results when implementing TAVR across diverse patient subgroups \citep{khan2025comparison}, the long-term outcomes of TAVR remain incompletely understood. Several works have reported higher \citep{ahmed2024meta} or comparable \citep{caminiti2024long, talanas2024long, leon2025transcatheter, khan2020comparison} long-term mortality rates compared to SAVR, while TAVR has also been associated with higher rates of post-surgical pacemaker implantation \citep{talanas2024long, kazemian2024trends} but lower risks of stroke and bleeding \cite{talanas2024long, harvey2024trends}. Other studies have examined the relative benefits of TAVR and SAVR across specific subgroups, such as younger patients \citep{jorgensen2025three}, those with chronic obstructive pulmonary disease (COPD) \citep{khan2021outcomes}, patients with low-flow, low-gradient stenosis \citep{ueyama2021impact}, and patients with specific anatomical features \citep{ayyad2024efficacy, amin2025evaluating, rodes2024transcatheter}.

The large number of studies and meta-analyses, often using different endpoints, methodologies, and yielding conflicting conclusions, has resulted in persistent heterogeneity in real-world treatment selection. Decisions remain shaped by institutional preferences, operator experience, and patient-specific anatomy or physiology. As a result, outcomes vary across hospitals and subgroups, and clinical guidelines become blurred, highlighting the need for data-driven frameworks that go beyond meta-analyses toward individualized prescription.

Most prior work has focused on predictive models estimating post-procedural mortality or complication risk for SAVR \citep{o2018society, vassiliou2021novel} and TAVR \citep{gupta2021predicting, yamamoto2021clinical, bruggemann2024predicting, codner2018mortality}. While useful for risk stratification, these models are not prescriptive; they do not recommend which treatment a given patient should receive. Beyond guideline-based recommendations, few studies offer systematic, data-driven tools to support SAVR vs. TAVR decision-making. Notably, manually derived prescriptive decision trees combining clinical insights with CTA features have been proposed \citep{chen2025lifetime}, but such rule-based approaches are difficult to scale and may be inconsistent across populations. More recent work has shown that patient matching and counterfactual modeling can mitigate confounding in observational cohorts, providing a promising path for precision treatment selection \citep{bertsimas2024road}.

Building on this foundation, we propose an interpretable prescriptive framework that integrates data from multiple sources to recommend the optimal treatment (TAVR or SAVR) for low to intermediate risk patients with severe aortic stenosis. The approach combines prognostic matching \citep{bertsimas2024road}, counterfactual outcome modeling, and an interpretable Optimal Policy Tree (OPT) \citep{amram2022optimal} trained on counterfactual predictions. We first create a balanced cohort via prognostic matching to approximate randomization and train counterfactual models to estimate each patient’s 5-year mortality risk under TAVR and SAVR. The OPT then groups patients into clinically meaningful subgroups and prescribes the treatment that minimizes long-term mortality within each group.

We evaluate the framework using data from two independent medical centers: Hartford Hospital for model development and internal validation, and St. Vincent’s Hospital for external testing, both located in Connecticut, USA. In both datasets, the learned policy shows improved estimated 5-year mortality relative to current practice and baseline strategies (random, all-SAVR, all-TAVR), while maintaining strong concordance with real-world decisions and alignment with established clinical insights.

By integrating optimization, prescriptive analytics, and interpretable modeling, this work provides, to our knowledge, the first transparent, machine-learning–based decision support tool for aortic valve replacement planning. Validated across internal and external cohorts, it lays the groundwork for precision medicine systems that guide individualized, evidence-based selection between TAVR and SAVR.

% \subsection{Related Work}

% \subsection{Contributions}

% This study makes three main contributions to the growing field of interpretable prescriptive modeling in cardiovascular medicine:

% Interpretable prescriptive framework:
% We develop an Optimal Policy Tree that recommends between TAVR and SAVR based on multimodal clinical and anatomical data. The model combines causal inference with rule-based learning, yielding transparent, patient-specific treatment policies that clinicians can easily interpret.

% Robust causal design for observational data:
% To address the confounding inherent in real-world cohorts, we apply prognostic matching to emulate randomization and train counterfactual outcome estimators to evaluate long-term mortality under each treatment. This ensures that the learned policy is grounded in causal rather than purely associative evidence.

% Comprehensive validation across hospitals:
% We validate our framework internally (Hartford) and externally (St. Vincent’s), demonstrating consistent improvement in estimated 10-year mortality compared with real-world prescriptions, random assignments, and single-treatment baselines.
% The policy maintains high concordance with clinical practice, recovering established medical insights while quantifying patient-level benefit.

% Together, these contributions establish a reproducible, interpretable, and empirically validated pathway toward data-driven decision support for structural heart disease, bridging predictive analytics and precision prescription.

% -----------------------------
\begin{center}
\fbox{
\begin{minipage}{0.46\textwidth}
\section*{Research in context}

\noindent\textbf{Evidence before this study}

Before undertaking this work, we reviewed the extensive literature comparing SAVR and TAVR in low and intermediate risk patients, including randomized trials, observational registries, guideline documents, and outcome-prediction models. Although numerous trials and meta-analyses have evaluated TAVR versus SAVR across different endpoints, populations, and follow-up windows, their conclusions are heterogeneous and often non-actionable for individual treatment decisions. Existing models primarily predict mortality or complications after one procedure rather than estimating which treatment would yield the best long-term outcome for a specific patient. Manually developed rule-based or subgroup-driven decision tools have limited generalizability due to inconsistent endpoints and population differences. We found no interpretable, data-driven framework that integrates real-world data, confounding mitigation, and individualized counterfactual outcomes to guide personalized treatment selection between SAVR and TAVR.

\noindent\textbf{Added value of this study}

This study introduces an interpretable, prescriptive framework for individualized treatment selection between SAVR and TAVR. By integrating prognostic matching, counterfactual outcome modeling, and an Optimal Policy Tree, the approach moves beyond risk prediction toward treatment recommendations tailored to each patient’s estimated long-term outcomes. The model yields transparent, clinically coherent rules and demonstrates consistent improvement in estimated outcomes across both internal and external real-world cohorts.

\noindent\textbf{Implications of all the available evidence}

Existing evidence shows that the relative benefits of SAVR and TAVR vary with anatomy, frailty, and comorbidities. When combined with these findings, our results highlight the potential for data-driven, individualized decision-support tools to improve treatment planning for severe aortic stenosis. Prospective validation and extension to additional outcomes, such as complications and durability, are important next steps toward clinical translation.
\end{minipage}
}
\end{center}

\section{Methods}

\noindent
This section outlines the methodology employed for our prescriptive framework, including the data collection process, the matching procedure used to create a more trial-like cohort, and the development of the final prescriptive model.

\subsection{Data Sources} \label{subsec: data_sources}
\noindent
We collected data from the Hartford HealthCare system, one of the largest hospital networks in Connecticut. Hartford Hospital served as the training cohort, and St. Vincent’s Hospital as the external validation cohort, with consistent data collection across institutions.

To mirror the information available to clinicians at the time of treatment decision-making, we integrated patient-level data from demographics, medical history, echocardiography, and computed tomography angiography (CTA).

The study cohort comprised low to intermediate risk patients (STS score $<$3·5) with severe aortic stenosis who underwent TAVR or SAVR between September 2016 and June 2023. We excluded patients with prior aortic interventions (e.g., CABG, prior TAVR/SAVR, or other valve procedures), concomitant procedures, or aortic regurgitation unrelated to aortic stenosis.

Patient information was sourced from two national registries: SAVR features from the Society of Thoracic Surgeons (STS) Adult Cardiac Surgery Database and TAVR features from the Transcatheter Valve Therapy (TVT) Registry. To our knowledge, this is the first dataset integrating both registries with echocardiographic and CTA-derived features for prescriptive modeling of valve replacement.

Long-term outcomes are difficult to track in observational data because patients may not return to the same hospital. To obtain accurate mortality information, we linked patients' records to the Connecticut Death Registry, by matching names and dates of birth. Although this approach does not capture out-of-state deaths, it provides a close proxy for complete mortality follow-up within Connecticut and covers all reported deaths from January 2015 to May 2024, consistent with the clinical and imaging data availability.

\subsection{Prescriptive Pipeline} \label{subsec: presc}

\noindent
Our methodology consists of four main components:
\begin{enumerate}
    \item Prognostic matching, which mitigates observed confounding using both patient covariates and outcomes~\citep{bertsimas2024road};
    \item Counterfactual estimation, which trains separate survival models for each treatment arm to estimate patient-level counterfactual outcomes;
    \item Policy learning, which fits an Optimal Policy Tree (OPT) that partitions patients into subgroups and prescribes one of the treatments for each subgroup; and
    \item Sample weighting, which adjusts for residual confounding (both observed and unobserved) and ensures that the final prescriptive model aligns with real-world clinical practice.
\end{enumerate}

The following subsections describe each step in detail.

\subsubsection{Problem Setting}

\noindent
Assuming that our dataset consists of $N$ patient observations, each patient $i$ is characterized by:
\begin{itemize}
    \item Covariates $\mathbf{x}_i \in \mathbb{R}^p$, where $p$ is the number of features. These include demographics, comorbidities, prior procedures, and imaging-derived measurements (echocardiography and CTA);
    \item Treatment assignment $t_i \in T$, where $T = \{\text{SAVR}, \text{TAVR}\}$, representing the treatment each patient received; and
    \item Outcome $y_i \in \{0,1\}$, indicating whether the patient died ($1$) or survived ($0$) within a 5-year horizon after the procedure.
\end{itemize}

The objective of the prescriptive model is to learn a treatment rule
\[
\tau(\mathbf{x}) : \mathbb{R}^p \rightarrow \mathcal{T},
\]
that assigns each patient to the treatment expected to minimize their probability of 5-year mortality. 

\subsection{Prognostic Matching} \label{subsec: prog_match}

\noindent
As with all observational data, our raw dataset is subject to confounding. Before estimating treatment effects or deriving optimal policies, we first reduce bias through prognostic matching. 

Prognostic matching \citep{bertsimas2024road} extends traditional covariate matching by also incorporating proximity in predicted outcomes, yielding patient pairs that are similar both in their baseline features and expected prognosis. 

We first stratified patients by their STS Risk Score, a validated measure of frailty and comorbidity presence. Within each risk stratum, we then performed 1:1 nearest-neighbor matching between SAVR and TAVR patients using a distance metric defined on their clinical covariates. This produced a balanced cohort in which observed confounding was substantially mitigated, approximating a trial-like population.

\subsection{Counterfactual Estimation} \label{subsec: ce}

Following prognostic matching, we trained two Survival Random Forest models, one for each treatment arm, to estimate the 5-year mortality risk conditional on patient characteristics. Each model was trained exclusively on patients who actually received the corresponding treatment, with cross-validation used to prevent overfitting and to ensure generalizable survival predictions. The use of survival analysis was essential, as follow-up durations varied substantially across patients, and proper handling of right-censoring was necessary to avoid biased risk estimation.

We then used these models to estimate counterfactual outcomes for all patients, i.e. the predicted mortality risk under both SAVR and TAVR. Training separate models per treatment empirically yielded larger differences between the two predicted risks compared with a single joint model (with a treatment indicator). This separation provides the Optimal Policy Tree with clearer treatment-effect heterogeneity and enables more informative splits.

\subsection{Optimal Policy Tree} \label{subsec: opt}

\noindent
The final step of the pipeline learns a prescriptive model that recommends a treatment for each patient based on their features and predicted counterfactuals. 

We employed Optimal Policy Trees (OPT) due to their interpretability and suitability for clinical settings. Each leaf of the tree corresponds to a subgroup of patients with similar characteristics and a recommended treatment. The OPT minimizes the following objective function, which corresponds to the expected average outcome (5-year mortality):

\begin{equation}\label{eq:opt_objective}
\min_{\tau(\cdot)} \sum_{i=1}^{N} \sum_{t \in T} \mathds{1}\{\tau(\mathbf{x}_i) = t\} \cdot \Gamma_{i,t},
\end{equation}
where $\Gamma_{i,t}$ denotes the predicted 5-year mortality risk for patient $i$ under treatment $t$, estimated from the survival models.

\subsubsection{Sample Weighting} 

\noindent
Even after matching, residual differences may persist between the treatment groups due to remaining observed or unobserved confounding. This can often be observed as persistent differences in average outcomes between arms, beyond what can be attributed to treatment effects alone. 

To address this, we adapted a sample-weighting approach \citep{bertsimas2024road}. The goal is to align outcome expectations across treatments, in order to train a prescriptive model that is both fair and clinically realistic.

Let 
\begin{align*}
\mathcal{T_S} &= \text{patients who received SAVR},\\
\mathcal{T_T} &= \text{patients who received TAVR},\\
\mathcal{G} &= \text{patients with a good outcome (survival)},\\
\mathcal{B} &= \text{patients with a bad outcome (mortality)}.
\end{align*}

We first identify the treatment
\[
t^* = \arg\min_{t \in T} \sum_i \Gamma_{i,t},
\]
that corresponds to the lowest average predicted mortality across all patients. 
We then apply a uniform weight $w$ to the subset of patients
\begin{equation}
    i \in (\mathcal{T}_{t^*} \cap \mathcal{B}) \cup (\mathcal{T}_{t} \cap \mathcal{G}), \quad t \neq t^*.
\end{equation}

In other words, we upweight:
\begin{itemize}
    \item patients who received the ``better'' treatment ($t^*$) but had a bad outcome; and
    \item patients who received the ``worse'' treatment ($t \neq t^*$) but had a good outcome.
\end{itemize}

This weighting scheme prevents the prognostic models from becoming overly optimistic for one treatment and overly pessimistic for the other, leading to more balanced counterfactual estimates across treatments.

We iteratively increased $w$, re-estimated the counterfactuals, and retrained the policy tree. The final tree was selected to jointly maximize sensitivity, specificity, and clinical interpretability, favoring models that revealed meaningful heterogeneity across demographic, clinical, and imaging features.

We define:
\begin{equation}
\text{Sensitivity} = \frac{|\mathcal{B} \cap \{i : \hat{t}_i \neq t_i\}|}{|\mathcal{B}|},
\end{equation}
and
\begin{equation}
\text{Specificity} = \frac{|\mathcal{G} \cap \{i : \hat{t}_i = t_i\}|}{|\mathcal{G}|},
\end{equation}
where $t_i$ is the treatment actually received and $\hat{t}_i$ is the treatment prescribed by our model. In other words, sensitivity refers to the percentage of patients who had a bad outcome, and for which we prescribed a different treatment, with a potentially better outcome, and specificity to the percentage of patients who had a good outcome, and for whom we prescribed the same treatment as in real life.

% -----------------------------
\section{Results}

\noindent
In this section, we present the results of our experiments across the different stages of our pipeline. Section \ref{subsec: res_cohort_ch} discusses our initial cohort characteristics, while Section \ref{subsec:prog_matching_results} presents the effect of prognostic matching on our cohort. A thorough analysis of our prescriptive model, both on the Hartford Hospital cohort and the external St. Vincent's cohort, is presented in Sections \ref{subsec: res_opt} and \ref{subsec: eval_metrics}.

\subsection{Cohort Characteristics} \label{subsec: res_cohort_ch}

\noindent
Our initial Hartford cohort included 1,352 patients, but CTA measurements were missing for most; because these anatomical features are clinically important for treatment selection, we restricted the analysis to 372 patients with available CTA data. The final cohort included 151 SAVR and 221 TAVR patients. Most variables had minimal missingness aside from two echocardiographic measurements (aortic valve mean gradient and velocity–time integral), which we kept because of their clinical relevance. Remaining missing values were imputed using k-nearest neighbors (KNN). Baseline characteristics that differed significantly between groups prior to matching aligned with clinical expectations, with TAVR patients generally older and higher risk. Full covariate lists, missingness, and statistical comparisons appear in \appSubsecHartford. 

% \begin{table}[H]
% \centering
% \footnotesize
% \caption{Selected baseline characteristics with significant differences between TAVR and SAVR in the Hartford Hospital cohort (pre-matching). Values are mean$\pm$SD.}
% \label{tab:baseline_main}
% \begin{tabular}{lcc}
% \toprule
% \textbf{Variable} & \textbf{TAVR} & \textbf{SAVR} \\
% \midrule
% Age (years) & 78.9$\pm$7.1 & 70.8$\pm$7.9 \\
% STS Predicted Risk of Mortality (\%) & 2.13$\pm$0.78 & 1.46$\pm$0.64 \\
% Prior PCI & 0.145$\pm$0.353 & 0.046$\pm$0.211 \\
% Prior MI & 0.095$\pm$0.294 & 0.033$\pm$0.180 \\
% Recent heart failure (within 2~weeks) & 0.434$\pm$0.497 & 0.007$\pm$0.081 \\
% Sinotubular junction diameter (mm) & 28.9$\pm$3.5 & 29.7$\pm$3.9 \\
% \addlinespace[2pt]
% \multicolumn{3}{r}{\footnotesize All variables $p<0.05$ (Welch or $\chi^2$ test).}\\
% \bottomrule
% \end{tabular}
% \vspace{-0.5em}
% \end{table}

For external validation, we used the cohort from St.~Vincent’s Hospital, comprising 36 SAVR and 269 TAVR patients. CTA measurements were not available in this dataset; therefore, these variables were imputed using the same KNN models fitted on the Hartford training data.  
Pre-matching differences between groups, along with complete feature summaries, are provided in \appSubsecStVinc.  

% \begin{table}[H]
% \centering
% \footnotesize
% \caption{Selected baseline characteristics with significant differences between TAVR and SAVR in the St.~Vincent’s Hospital cohort (pre-matching). Values are mean$\pm$SD.}
% \label{tab:sv_sig}
% \begin{tabular}{lcc}
% \toprule
% \textbf{Variable} & \textbf{TAVR} & \textbf{SAVR} \\
% \midrule
% Age (years) & 77.0$\pm$7.9 & 62.6$\pm$10.5 \\
% STS Predicted Risk of Mortality (\%) & 2.15$\pm$0.81 & 1.13$\pm$0.72 \\
% Porcelain aorta & 0.01$\pm$0.12 & 0.00$\pm$0.00 \\
% Peripheral arterial disease (history) & 0.09$\pm$0.29 & 0.00$\pm$0.00 \\
% Coronary artery disease (history) & 0.14$\pm$0.35 & 0.03$\pm$0.17 \\
% Proximal LAD disease & 0.08$\pm$0.27 & 0.00$\pm$0.00 \\
% Recent heart failure (2 wk) & 0.70$\pm$0.46 & 0.00$\pm$0.00 \\
% Aortic valve peak velocity (m/s) & 3.89$\pm$0.70 & 4.24$\pm$0.63 \\
% Aortic valve VTI (cm) & 97.3$\pm$13.2 & 85.7$\pm$17.3 \\
% \addlinespace[2pt]
% \multicolumn{3}{r}{\footnotesize All variables $p<0.05$ (Welch or $\chi^2$ test).}\\
% \bottomrule
% \end{tabular}
% \vspace{-0.6em}
% \end{table}

\begin{figure*}[t]
\centering
\subfloat[Hartford Hospital (training cohort)]{%
  \includegraphics[width=0.48\textwidth]{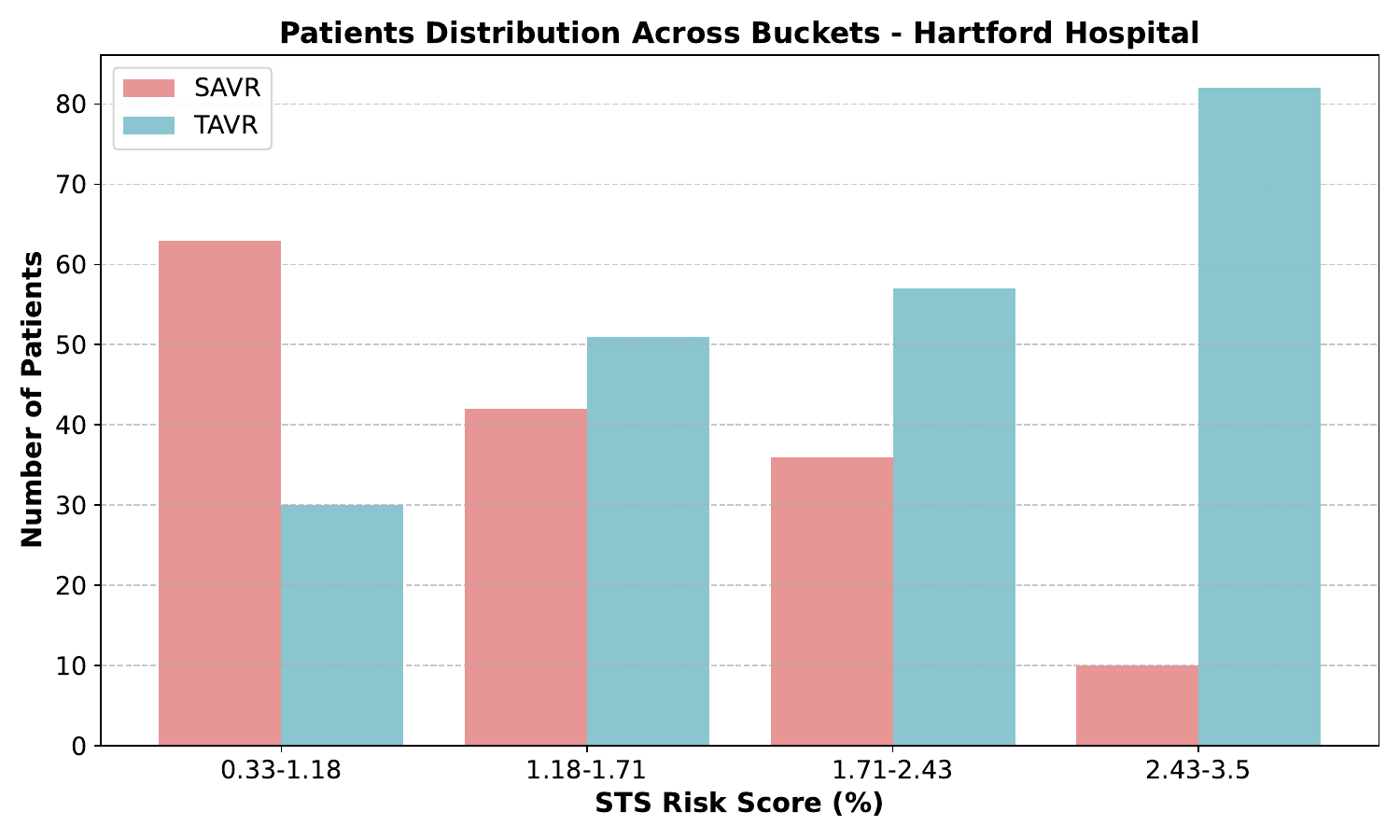}}
\hfill
\subfloat[St.~Vincent’s Hospital (external validation cohort)]{%
  \includegraphics[width=0.48\textwidth]{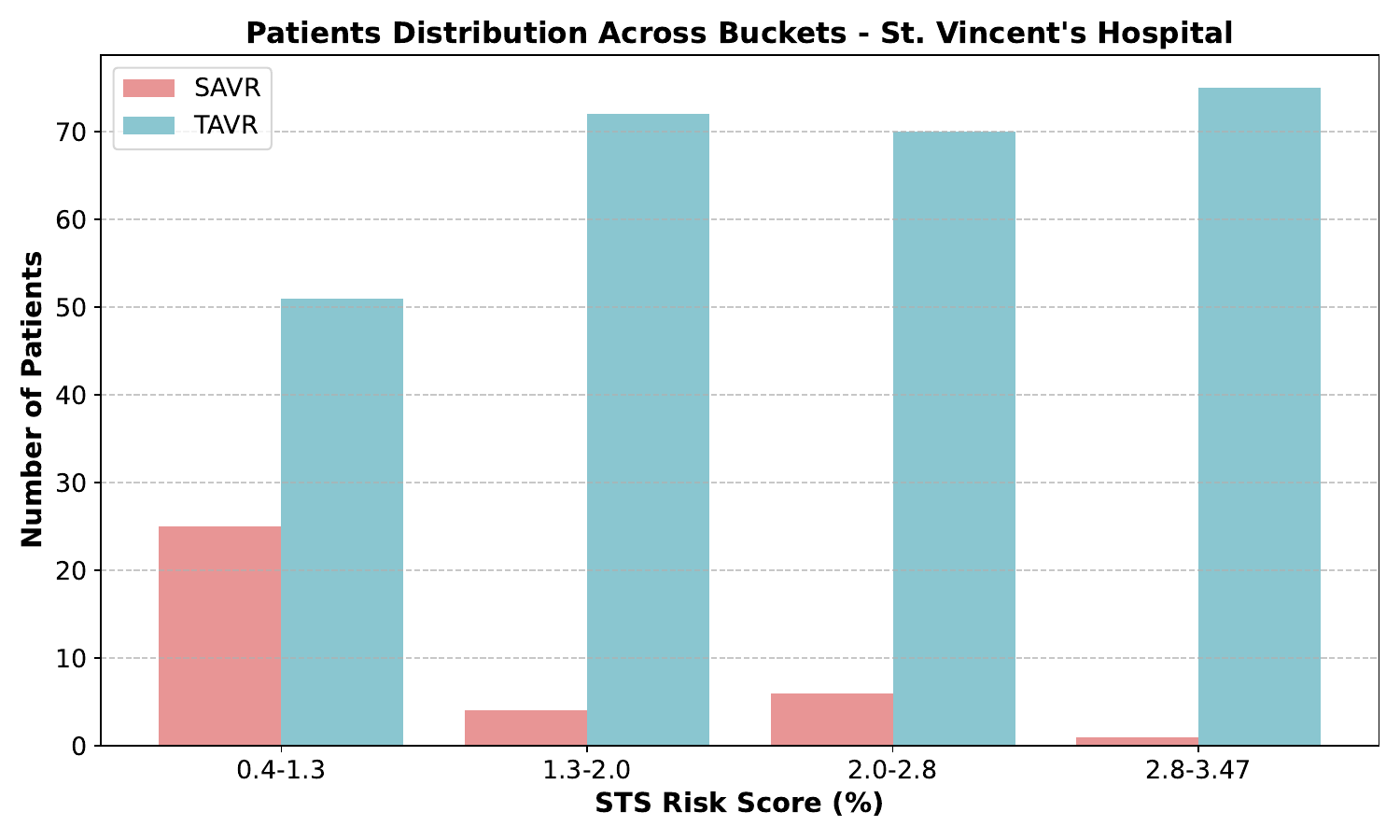}}
\caption{
Distribution of patients across Society of Thoracic Surgeons (STS) risk score strata in the Hartford (training) and St.~Vincent’s (external validation) cohorts. 
Both cohorts exhibit treatment imbalance, with TAVR concentrated in higher-risk strata, motivating the use of prognostic matching on the training data to ensure comparability.}
\label{fig:sts_dual}
\vspace{-0.5em}
\end{figure*}

Consistent with the Hartford cohort, higher-risk patients at St.~Vincent’s were more frequently treated with TAVR, even within this overall lower-risk population. This pattern becomes more apparent when stratifying patients by STS Predicted Risk of Mortality (STS Risk Score).  
Figure~\ref{fig:sts_dual} illustrates the distribution of TAVR and SAVR patients across STS risk strata in both datasets, highlighting the clear treatment imbalance even within the low to intermediate risk range ($\text{STS}<$ 3·5\%). This is also reflected in the average historical mortality rate in a 5-year horizon, as presented in Table \ref{tab:mortality_combined}, Panel A, where TAVR patients consistently exhibit higher long-term mortality than SAVR patients.

% \begin{table}[H]
% \centering
% \footnotesize
% \caption{Observed 5-year-horizon mortality rates across cohorts and treatment groups. Values represent the proportion of patients deceased up to five years after their surgery.}
% \label{tab:mortality_rates}
% \begin{tabular}{lccc}
% \toprule
% \textbf{Cohort} & \textbf{Overall (\%)} & \textbf{SAVR (\%)} & \textbf{TAVR (\%)} \\
% \midrule
% Hartford Hospital & 9·68 & 5·30 & 12·67 \\
% St.~Vincent’s Hospital & 13·11 & 5·56 & 14·13 \\
% \bottomrule
% \end{tabular}
% \vspace{-0.5em}
% \end{table}

These observations underscore the importance of prognostic matching in the Hartford training cohort to achieve balanced treatment groups across the risk spectrum, before proceeding with policy learning.

\subsection{Prognostic Matching} \label{subsec:prog_matching_results}
\noindent 
To address the observed imbalance, we applied prognostic matching within the Hartford (training) cohort by pairing SAVR and TAVR patients within each STS risk bucket. This procedure resulted in a cohort of 236 patients that is slightly shifted toward lower-risk patients, as presented in Figure \appFigDistPostMatching. 
The resulting improvement in covariate balance is summarized in Figure~\appFigLovePlotRest, which displays the absolute standardized mean differences (SMDs) before and after matching for all variables that were different with statistical significance prior to matching.

All selected covariates showed reduced imbalance after matching, yielding a more comparable cohort across treatment groups. The full list of covariates, statistical comparisons, and SMDs, along with more details on how SMDs are calculated, are reported in \appSeProgMatching. Of the 39 features, 22 had lower or equal SMDs post-matching. Although not all covariates are perfectly aligned, balancing on the STS Risk Score ensures alignment on the key prognostic dimension for 5-year mortality and captures joint effects across features.

The impact of prognostic matching is also evident when comparing mortality rates before and after matching (Table~\ref{tab:mortality_combined}, Panel B). The residual mortality difference between SAVR and TAVR after matching likely reflects a combination of small residual observed confounding, unobserved confounding, and genuine treatment effect heterogeneity. The subsequent sample weighting step aims to further mitigate the confounding effect on the differences.

% \begin{table}[H]
% \centering
% \footnotesize
% \caption{Hartford Hospital: observed 5-year mortality before and after prognostic 1:1 matching (Pre-match and Post-match respectively). Values are proportions of patients deceased by 5 years.}
% \label{tab:hartford_mortality_pre_post}
% \begin{tabular}{lcc}
% \toprule
% \textbf{Group} & \textbf{Pre-match (\%)} & \textbf{Post-match (\%)} \\
% \midrule
% Overall & 9·68 & 7·63 \\
% SAVR    & 5·30 & 6·78 \\
% TAVR    & 12·67 & 8·47 \\
% \bottomrule
% \end{tabular}
% \vspace{-0.5em}
% \end{table}

\begin{table}[H]
\centering
\footnotesize
\caption{Observed 5-year mortality rates. 
Panel~A reports mortality across cohorts and treatment groups. 
Panel~B reports Hartford Hospital mortality before and after prognostic 1:1 matching. 
All values represent the proportion of patients deceased by 5 years.}
\label{tab:mortality_combined}
\setlength{\tabcolsep}{5pt}
\begin{tabular}{lccc}
\toprule
\multicolumn{4}{l}{\textbf{Panel A. Observed 5-year mortality across cohorts}} \\
\midrule
\textbf{Cohort} & \textbf{Overall (\%)} & \textbf{SAVR (\%)} & \textbf{TAVR (\%)} \\
\midrule
Hartford Hospital      & 9·68  & 5·30 & 12·67 \\
St.~Vincent’s Hospital & 13·11 & 5·56 & 14·13 \\
\midrule
\\[-0.7em]
\multicolumn{4}{l}{\textbf{Panel B. Hartford Hospital: pre- and post-match mortality}} \\
\midrule
\textbf{Group} & \textbf{Pre-match (\%)} & \textbf{Post-match (\%)} & \\
\midrule
Overall & 9·68  & 7·63 & \\
SAVR    & 5·30  & 6·78 & \\
TAVR    & 12·67 & 8·47 & \\
\bottomrule
\end{tabular}
\vspace{-0.5em}
\end{table}

\subsection{Optimal Policy Tree Prescriptions} \label{subsec: res_opt}

\noindent
After constructing a balanced training cohort through prognostic matching, we evaluated the prescriptive performance of our optimal policy model.
The analysis proceeded in two stages: (1) training counterfactual outcome estimators on the matched Hartford cohort, and (2) learning an interpretable optimal policy tree that recommends the preferred treatment (TAVR or \\SAVR) for each patient based on their features.
We then evaluated both the internal (Hartford) and external (St.~Vincent’s) validity of the learned policy.

\begin{figure}[t]
\centering
\includegraphics[width=0.45\textwidth]{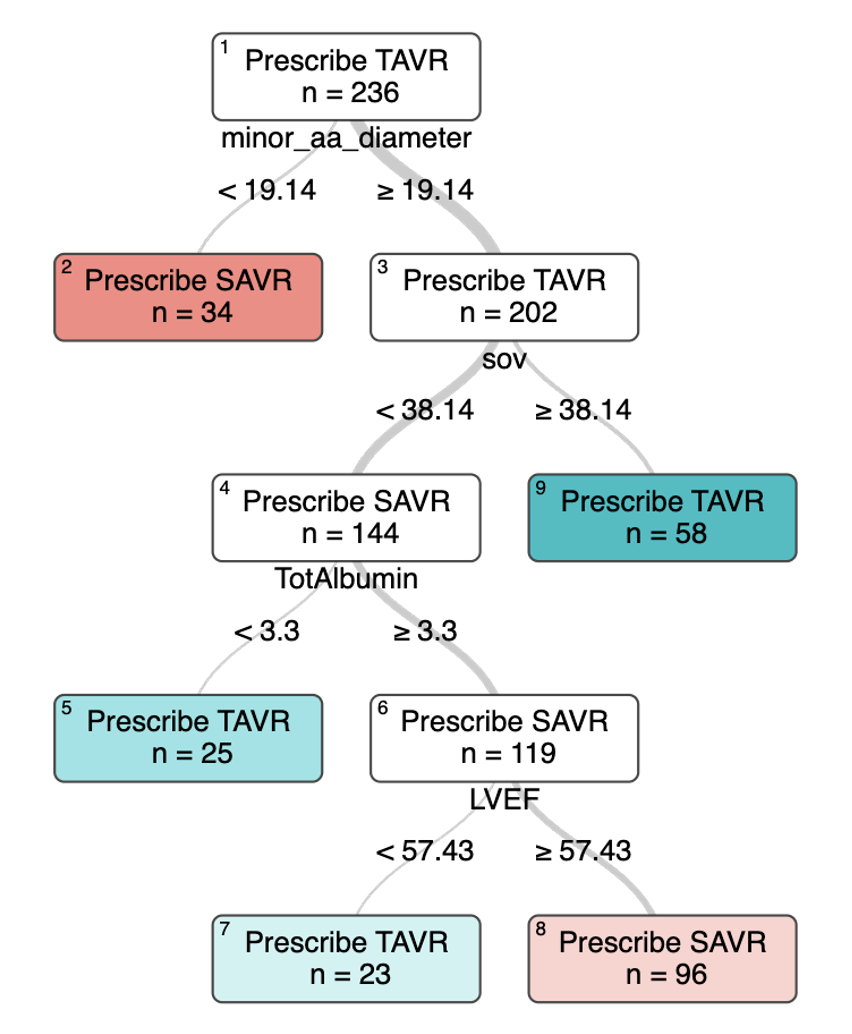}
\caption{
Final Optimal Policy Tree selected after tuning the sample weight $w$.
Splitting features include CTA measurements (minor aortic annulus diameter, sinus of Valsalva diameter (mm)), echocardiographic measurements (Left Ventricular Ejection Fraction (LVEF)), and the patients' total albumin, which is a common frailty indicator.
}
\label{fig:opt}
\vspace{-0.5em}
\end{figure}

Among multiple candidate trees trained with varying complexity parameters and sample-weight configurations, we selected the one achieving the highest average of sensitivity and specificity in the Hartford cohort, while also maintaining clinical plausibility and treatment balance.
This tree, presented in Figure \ref{fig:opt}, corresponds to the configuration with a sample weight of $w=$1·8, which upweights SAVR patients with poor outcomes and TAVR patients with favorable outcomes. More details on the effect of sample weighting on the sensitivity and specificity of the resulting trees can be found in \appSecSampleWeight. Notably, the resulting tree is both compact and clinically interpretable, partitioning the cohort into five subgroups with distinct characteristics, combining simplicity with individualized treatment recommendations.

Table~\ref{tab:optree_nodes_combined} presents a leaf-level summary of the selected tree, reporting for each subgroup the number (\%) of patients who received each treatment and their observed 5-year mortality. Across all five leaves, the prescribed treatment always corresponds to the option with the lowest observed mortality, indicating strong alignment between the model’s recommendations and real-world outcomes in the matched Hartford cohort. The prescribed treatments are also clinically plausible: in every node, the recommended option is well represented among real-world decisions, suggesting consistency with established practice patterns.

When applied to the St.~Vincent’s Hospital cohort, the policy produced directionally consistent patterns across all but one leaf (Node 7). This exception is likely due to the strong imbalance in the St.~Vincent’s dataset, with far more TAVR than SAVR patients, and higher average STS risk scores among TAVR cases; however it is evident that in this subgroup TAVR is prescribed considerably more than SAVR in real life, making the prescription aligned with current clinical practice. Overall, these results highlight the robustness and generalizability of the policy, even when evaluated on a cohort with substantially different treatment distributions and risk profiles, since prognostic matching was only performed on the Hartford dataset.

% \begin{table}[H]
% \centering
% \footnotesize
% \caption{
% St.~Vincent’s cohort: leaf-level summary of the Optimal Policy Tree applied externally.
% Each node represents a subgroup of patients for whom the model prescribes either SAVR or TAVR.
% For each subgroup, we report the number (\%) of patients who actually received each treatment and their observed 5-year mortality (proportion).}
% \label{tab:optree_nodes_stvinc}
% \setlength{\tabcolsep}{2pt}
% \begin{tabular}{lcccccc}
% \toprule
% \multirow{2}{*}{\textbf{Node}} & \multirow{2}{*}{\textbf{Prescribed}} &
% \multicolumn{2}{c}{\shortstack{\textbf{SAVR}\\\textbf{patients}}} &
% \multicolumn{2}{c}{\shortstack{\textbf{TAVR}\\\textbf{patients}}} \\
% \cmidrule(lr){3-4}\cmidrule(lr){5-6}
%  &  & \textbf{N (\%)} & \textbf{Mortality (\%)} & \textbf{N (\%)} & \textbf{Mortality (\%)} \\
% \midrule
% 2  & SAVR & 1 (2·8\%)  & 0·0 & 9 (3·4\%)   & 11·1 \\
% 5  & TAVR & 2 (5·6\%)  & 0·0 & 1 (0·4\%)   & 0·0 \\
% 7  & TAVR & 2 (5·6\%)  & 0·0 & 60 (22·3\%) & 23·3 \\
% 8  & SAVR & 23 (63·9\%) & 4·3 & 165 (61·3\%) & 12·1 \\
% 9  & TAVR & 8 (22·2\%) & 12·5 & 34 (12·6\%)  & 8·8 \\
% \bottomrule
% \end{tabular}
% \vspace{-0.5em}
% \end{table}

\begin{table*}[t]
\centering
\footnotesize
\caption{
Leaf-level summary of the Optimal Policy Tree for both cohorts.
For each node, the table reports the number (\%) of patients who actually received each treatment
and their observed 5-year mortality (proportion).}
\label{tab:optree_nodes_combined}
\setlength{\tabcolsep}{2pt}
\begin{tabular}{l l c c c c c}
\toprule
\multirow{2}{*}{\textbf{Cohort}} &
\multirow{2}{*}{\textbf{Node}} &
\multirow{2}{*}{\textbf{Prescribed}} &
\multicolumn{2}{c}{\shortstack{\textbf{SAVR}\\\textbf{patients}}} &
\multicolumn{2}{c}{\shortstack{\textbf{TAVR}\\\textbf{patients}}} \\
\cmidrule(lr){4-5}\cmidrule(lr){6-7}
 &  &  & \textbf{N (\%)} & \textbf{Mortality (\%)} &
     \textbf{N (\%)} & \textbf{Mortality (\%)} \\
\midrule
\textbf{Hartford} & 2  & SAVR & 18 (15·3\%) & 0·0 & 16 (13·6\%) & 12·5 \\
                  & 5  & TAVR & 16 (13·6\%) & 6·3 & 9 (7·6\%)   & 0·0  \\
                  & 7  & TAVR & 9 (7·6\%)   & 11·1 & 14 (11·9\%) & 0·0  \\
                  & 8  & SAVR & 47 (39·8\%) & 2·1 & 49 (41·5\%) & 10·2 \\
                  & 9  & TAVR & 28 (23·7\%) & 17·9 & 30 (25·4\%) & 10·0 \\
\midrule
\textbf{St.~Vincent's} & 2  & SAVR & 1 (2·8\%)  & 0·0 & 9 (3·4\%)    & 11·1 \\
                       & 5  & TAVR & 2 (5·6\%)  & 0·0 & 1 (0·4\%)    & 0·0  \\
                       & 7  & TAVR & 2 (5·6\%)  & 0·0 & 60 (22·3\%)  & 23·3 \\
                       & 8  & SAVR & 23 (63·9\%)& 4·3 & 165 (61·3\%) & 12·1 \\
                       & 9  & TAVR & 8 (22·2\%) & 12·5 & 34 (12·6\%) & 8·8  \\
\bottomrule
\end{tabular}
\vspace{-0.5em}
\end{table*}

\subsection{Evaluation Metrics} \label{subsec: eval_metrics}

\noindent
As discussed above, the selected Optimal Policy Tree achieved the best balance of sensitivity and specificity among all candidate trees. Here, sensitivity denotes the proportion of patients with adverse outcomes whose prescribed treatment was changed by the model, while specificity reflects the proportion of patients with good outcomes whose treatment remained unchanged. High sensitivity identifies patients for whom an alternative treatment may have improved outcomes, whereas high specificity indicates preservation of successful real-life prescriptions.

Table~\ref{tab:policy_combined}, Panel A reports the performance metrics and concordance with real-life decisions for both cohorts. In the Hartford cohort, the tree attains high sensitivity and moderate-to-high specificity, with 50\% of recommendations matching real-life practice. Sensitivity and specificity are lower in the external St.~Vincent’s cohort due to domain differences, yet the model still updates approximately 55\% of poor-outcome prescriptions and maintains 40\% concordance with real-life prescriptions, indicating reasonable generalization.

% \begin{table}[H]
% \centering
% \footnotesize
% \caption{Policy discrimination metrics with 95\% bootstrap CIs.}
% \label{tab:sens_spec_metrics_ci}
% \setlength{\tabcolsep}{3.5pt} % tighter spacing
% \begin{tabular}{lccc}
% \toprule
% \textbf{Cohort} & \textbf{Sens. [95\% CI]} & \textbf{Spec. [95\% CI]} & \textbf{Conc. [95\% CI]} \\
% \midrule
% Hartford (train)     & 0·78 [0·56--0·95] & 0·52 [0·45--0·59] & 0·50 [0·44--0·57] \\
% St.~Vincent’s (ext.) & 0·55 [0·39--0·69] & 0·38 [0·32--0·44] & 0·39 [0·33--0·44] \\
% \bottomrule
% \end{tabular}
% \vspace{-0.5em}
% \end{table}

% \noindent
To assess clinical utility, we compared expected 5-year mortality under current practice with four alternative strategies: random assignment, all-SAVR, all-TAVR, and our policy tree, using counterfactual mortality models trained on the matched Hartford cohort. Table~\ref{tab:policy_combined}, Panel B shows that our policy improves estimated mortality by more than 10\% in both datasets, despite being trained solely on Hartford data.

% \begin{table}[H]
% \centering
% \footnotesize
% \caption{Relative improvement (\%) in estimated 5-year mortality versus real-life treatment under different policies, with 95\% bootstrap CIs. Positive values denote improvement.}
% \label{tab:policy_value_combined_ci}
% \setlength{\tabcolsep}{3pt} % tighter for one-column fit
% \begin{tabular}{lcc}
% \toprule
% \textbf{Policy} & \textbf{Hartford} & \textbf{St.~Vincent’s} \\
% \midrule
% All SAVR             & --1·0 [--5·9, +3·2] & +4·6 [+0·7, +8·1] \\
% All TAVR             & +1·4 [--3·7, +6·5]  & +1·5 [+0·0, +3·1] \\
% Random Prescr.       & +0·1 [--4·4, +5·1]  & +3·1 [+0·0, +5·9] \\
% \textbf{Ours (OPT)}  & \textbf{+20·2 [+16·5, +23·5]} & \textbf{+13·8 [+11·5, +16·4]} \\
% \bottomrule
% \end{tabular}
% \vspace{-0.5em}
% \end{table}

We further validated these findings using a leaf-level analysis that relies only on observed outcomes. Following our prior TAVR methodology \citep{paschalidis2025optimalvalveprescriptiontranscatheter}, we assigned each patient either their observed outcome (if the model and real-life treatment agreed) or the average observed outcome of similar patients in the same leaf who historically received the recommended treatment. This model-free estimate provides a direct assessment within locally homogeneous subgroups.

% \begin{table}[H]
% \centering
% \footnotesize
% \caption{Leaf-analysis (empirical) relative improvement in 5-year mortality under the learned policy, with 95\% bootstrap CIs.}
% \label{tab:leaf_analysis_ci}
% \setlength{\tabcolsep}{10pt}
% \begin{tabular}{lc}
% \toprule
% \textbf{Cohort} & \textbf{Leaf Analysis Improvement} \\
% \midrule
% Hartford Hospital      & +55·45\% \;[+31·37\%,\; +75·68\%] \\
% St.~Vincent’s Hospital & +43·42\% \;[+27·51\%,\; +57·59\%] \\
% \bottomrule
% \end{tabular}
% \vspace{-0.5em}
% \end{table}

\begin{table*}[t]
\centering
\footnotesize
\caption{
Summary of policy performance across cohorts, with 95\% bootstrap confidence intervals.
Panel~A reports policy discrimination metrics. 
Panel~B reports the relative improvement (\%) in estimated 5-year mortality under different policies, where positive values denote improvement over real-world treatment. 
Panel~C reports the empirical leaf-analysis improvement under the learned policy.}
\label{tab:policy_combined}
\setlength{\tabcolsep}{2pt}
\begin{tabular}{lccc}
\toprule
\multicolumn{4}{l}{\textbf{Panel A. Policy discrimination metrics}} \\
\midrule
\textbf{Cohort} & \textbf{Sens. [95\% CI]} & \textbf{Spec. [95\% CI]} & \textbf{Conc. [95\% CI]} \\
\midrule
Hartford (train)     & 0·78 [0·56--0·95] & 0·52 [0·45--0·59] & 0·50 [0·44--0·57] \\
St.~Vincent’s (ext.) & 0·55 [0·39--0·69] & 0·38 [0·32--0·44] & 0·39 [0·33--0·44] \\
\midrule
\\[-0.7em]
\multicolumn{4}{l}{\textbf{Panel B. Relative improvement (\%) in estimated 5-year mortality}} \\
\midrule
\textbf{Policy} & \textbf{Hartford} & \textbf{St.~Vincent’s} & \\
\midrule
All SAVR             & --1·0 [--5·9, +3·2] & +4·6 [+0·7, +8·1] & \\
All TAVR             & +1·4 [--3·7, +6·5]  & +1·5 [+0·0, +3·1]  & \\
Random Prescr.       & +0·1 [--4·4, +5·1]  & +3·1 [+0·0, +5·9]  & \\
\textbf{Ours (OPT)}  & \textbf{+20·2 [+16·5, +23·5]} & \textbf{+13·8 [+11·5, +16·4]} & \\
\midrule
\\[-0.7em]
\multicolumn{4}{l}{\textbf{Panel C. Leaf-analysis empirical improvement}} \\
\midrule
\textbf{Cohort} & \textbf{Improvement [95\% CI]} & & \\
\midrule
Hartford Hospital      & +55·45\% \;[+31·37\%,\; +75·68\%] & & \\
St.~Vincent’s Hospital & +43·42\% \;[+27·51\%,\; +57·59\%] & & \\
\bottomrule
\end{tabular}
\vspace{-0.5em}
\end{table*}

The leaf-based analysis in Table \ref{tab:policy_combined}, Panel C indicates substantial reductions in mortality risk, exceeding 40\% in both cohorts. Even in the more imbalanced St.~Vincent’s dataset, the model yields large estimated benefits, suggesting that the learned rules capture stable and clinically coherent patterns. Overall, the policy tree meaningfully improves recommended treatment pathways while remaining interpretable and consistent with established medical reasoning.

% -----------------------------
\section{Discussion}

\noindent
In this study, we developed and validated an interpretable prescriptive model for individualized selection between surgical (SAVR) and transcatheter (TAVR) aortic valve replacement. Integrating prognostic matching, counterfactual outcome modeling, and an Optimal Policy Tree, the framework provides transparent, patient-level recommendations that are data-driven and clinically plausible. Results from the internal (Hartford) and external (St.~Vincent’s) cohorts consistently show improved estimated long-term outcomes over real-world practice and baseline strategies, while maintaining high interpretability and alignment with established clinical reasoning.

\subsection{Clinical Implications}

\noindent
The Optimal Policy Tree recovers clinically meaningful decision boundaries. Patients with smaller aortic annuli who undergo TAVR experience fewer bleeding complications than those treated with SAVR~\citep{ayyad2024efficacy}, but TAVR in this subgroup may carry higher long-term cardiovascular mortality~\citep{amin2025evaluating} or, in some studies, similar outcomes~\citep{rodes2024transcatheter}. SAVR is typically favored when the sinus of Valsalva is narrow due to the higher risk of coronary obstruction during TAVR, whereas larger sinuses make TAVR anatomically safer~\citep{arevalos2024coronary}. Within these anatomical groups, the model prescribes TAVR when total albumin is low (indicating frailty) or when LVEF is reduced - patterns supported by prior literature~\citep{attinger2021age,ludwig2023transcatheter}. Overall, the tree captures physiologically coherent relationships: TAVR is favored for frailer patients, those with lower LVEF, or those with larger sinus of Valsalva dimensons, whereas SAVR is preferred for smaller or surgically favorable anatomy.

Performance was strong across cohorts. In Hartford, the tree achieved high sensitivity (78\%) and moderate specificity (52\%), identifying patients with poor outcomes who may benefit from treatment reassignment while largely preserving favorable real-world prescriptions. Counterfactual evaluation showed estimated reductions in 5-year mortality of 20·2\% (Hartford) and 13·8\% (St.~Vincent’s). Leaf-based analysis further supported these findings, with relative improvements of 55\% and 43\%, respectively, when outcomes were approximated using observed results from similar patients in the same prescriptive subgroup. Importantly, recommendations remained clinically realistic, often matching the majority treatment within each leaf.

\subsection{Limitations}

\noindent
Several limitations should be acknowledged. Prognostic matching reduces but cannot fully eliminate selection bias; unmeasured confounders may still influence the training cohort. Mortality estimates rely on counterfactual survival models, which may exhibit calibration differences across institutions with distinct patient populations. The external cohort also had substantial missing CTA measurements, potentially limiting the evaluation of our model's generalization. Finally, our analysis focused solely on 5-year mortality, which was only captured inside Connecticut, and did not evaluate other clinically relevant endpoints such as complications, rehospitalization, or valve durability. Extending the framework to multi-objective settings and validating it prospectively across more institutions would strengthen its applicability.

\subsection{Contributions}

\noindent
This work is, to our knowledge, the first to apply prognostic matching for confounding mitigation prior to policy learning in observational cardiology datasets. It introduces the first interpretable, data-driven decision support tool that provides individualized treatment recommendations for SAVR vs.~TAVR in low to intermediate risk severe aortic stenosis, validated across two independent cohorts. By improving estimated outcomes while preserving interpretability and ease of deployment, the proposed framework offers a practical foundation for integrating prescriptive analytics into clinical decision-making for structural heart disease.

% -----------------------------
\section*{Contributors}
\noindent
All authors contributed to the study concept and overall research design. VS led the data curation, model development, statistical analyses, and implementation of all methodological components, and drafted the manuscript. MT, TA, JH, HH and RH provided domain expertise and supervised interpretation of findings. MT, TA, JH, HH, RH and DB critically revised the manuscript for important content. All authors approved the final version.

\section*{Declaration of interests}

\noindent
The authors declare no competing interests.

\section*{Data sharing}

\noindent
Patient data used in this study were obtained from participating institutions (Hartford Hospital and St. Vincent's Hospital) under data use agreements and cannot be publicly shared due to privacy restrictions. 

\section*{Acknowledgments}

\noindent
The authors thank the Hartford Hospital and St.~Vincent’s Hospital clinical teams for providing access to the patient data used in this study, and Hartford HealthCare for computational resources. The authors are also grateful to the clinicians who offered domain feedback during model interpretation and validation. The authors would like to warmly thank Stephanie Tran at MIT for her invaluable support and continuous assistance in the administration of this project.

\section*{Declaration of generative AI and AI-assisted technologies in the manuscript preparation process}
\noindent
During the preparation of this work the author(s) used ChatGPT in order to fix syntax and grammar errors. After using this tool/service, the author(s) reviewed and edited the content as needed and take(s) full responsibility for the content of the published article.

% -----------------------------
\bibliography{ref} 

\end{document}

\begin{appendices}
\renewcommand{\thesection}{\arabic{section}}

% \newpage

\mainheading{Supplementary Material}

\tableofcontents
\newpage

\section{Cohort characteristics} \label{app_sec: cohort_char}

\subsection{Hartford Hospital} \label{app_subsec: hartford}

\begin{table}[H]
\centering
\caption{Baseline characteristics of the Hartford Hospital cohort by treatment arm. Values are mean~$\pm$~SD, with overall missingness and $p$-values comparing TAVR and SAVR patients.}
\label{tab:baseline_cohorts}
\resizebox{\textwidth}{!}{
\begin{tabular}{lcccc}
\toprule
\textbf{Variable} & \textbf{TAVR (mean$\pm$SD)} & \textbf{SAVR (mean$\pm$SD)} & \textbf{$p$-value} & \textbf{Missing (\%)} \\
\midrule
\multicolumn{5}{l}{\textbf{Demographics}}\\
Sex (female = 1) & 0.33$\pm$0.47 & 0.40$\pm$0.49 & 0.178 & 0.00 \\
Age (years) & 78.91$\pm$7.06 & 70.81$\pm$7.85 & $<10^{-20}$ & 0.00 \\
Body Mass Index (kg/m$^2$) & 29.01$\pm$5.70 & 30.06$\pm$7.03 & 0.130 & 0.00 \\
STS Predicted Risk of Mortality (\%) & 2.13$\pm$0.78 & 1.46$\pm$0.64 & $<10^{-17}$ & 0.00 \\
\addlinespace
\multicolumn{5}{l}{\textbf{Prior Medical Conditions}}\\
Porcelain aorta & 0.00$\pm$0.00 & 0.00$\pm$0.00 & -- & 0.00 \\
Peripheral arterial disease (history) & 0.10$\pm$0.31 & 0.07$\pm$0.26 & 0.292 & 0.00 \\
Transient ischemic attack (history) & 0.04$\pm$0.19 & 0.04$\pm$0.20 & 0.862 & 0.00 \\
Prior myocardial infarction & 0.10$\pm$0.29 & 0.03$\pm$0.18 & 0.012 & 0.00 \\
Prior percutaneous coronary intervention & 0.14$\pm$0.35 & 0.05$\pm$0.21 & $<0.001$ & 0.00 \\
Home supplemental oxygen therapy & 0.01$\pm$0.09 & 0.03$\pm$0.16 & 0.233 & 0.00 \\
Current dialysis & 0.00$\pm$0.00 & 0.00$\pm$0.00 & -- & 0.00 \\
Hypertension (history) & 0.87$\pm$0.34 & 0.89$\pm$0.31 & 0.457 & 0.00 \\
Diabetes mellitus & 0.26$\pm$0.44 & 0.25$\pm$0.44 & 0.892 & 0.00 \\
Cerebrovascular disease (history) & 0.13$\pm$0.33 & 0.10$\pm$0.30 & 0.410 & 0.00 \\
Coronary artery disease (history) & 0.07$\pm$0.26 & 0.07$\pm$0.26 & 0.987 & 0.00 \\
Chronic obstructive pulmonary disease (history) & 0.25$\pm$0.43 & 0.25$\pm$0.43 & 0.952 & 0.00 \\
Immunosuppression (history) & 0.11$\pm$0.31 & 0.06$\pm$0.24 & 0.087 & 0.00 \\
Recent heart failure (within 2 weeks) & 0.43$\pm$0.50 & 0.01$\pm$0.08 & $<10^{-27}$ & 0.00 \\
Left main coronary disease & 0.01$\pm$0.09 & 0.00$\pm$0.00 & 0.158 & 0.00 \\
Proximal LAD disease & 0.01$\pm$0.09 & 0.00$\pm$0.00 & 0.158 & 0.00 \\
Intra-aortic balloon pump support & 0.00$\pm$0.00 & 0.00$\pm$0.00 & -- & 0.00 \\
Serum albumin (g/dL) & 4.10$\pm$0.37 & 4.17$\pm$0.40 & 0.175 & 22.85 \\
\addlinespace
\multicolumn{5}{l}{\textbf{CTA Measurements}}\\
Right coronary ostium height (mm) & 18.34$\pm$3.92 & 17.69$\pm$3.80 & 0.110 & 0.54 \\
Left coronary ostium height (mm) & 16.42$\pm$3.46 & 15.88$\pm$3.40 & 0.142 & 0.81 \\
Aortic root angulation (°) & 49.45$\pm$8.09 & 50.97$\pm$8.60 & 0.090 & 0.81 \\
Sinotubular junction diameter (mm) & 28.88$\pm$3.48 & 29.69$\pm$3.87 & 0.043 & 4.84 \\
Sinus of Valsalva diameter (mm) & 35.13$\pm$4.11 & 35.41$\pm$4.25 & 0.528 & 2.69 \\
Aortic annulus major diameter (mm) & 28.22$\pm$2.95 & 28.76$\pm$3.44 & 0.119 & 0.54 \\
Aortic annulus minor diameter (mm) & 22.37$\pm$2.69 & 22.71$\pm$2.83 & 0.245 & 0.81 \\
\addlinespace
\multicolumn{5}{l}{\textbf{Echocardiography Measurements}}\\
Left ventricular ejection fraction (\%) & 60.70$\pm$10.09 & 59.75$\pm$9.83 & 0.428 & 14.52 \\
Aortic valve mean gradient (mm Hg) & 45.59$\pm$13.20 & 43.11$\pm$11.58 & 0.112 & 18.55 \\
Aortic valve peak velocity (m/s) & 4.07$\pm$0.71 & 4.18$\pm$0.65 & 0.184 & 20.16 \\
Aortic insufficiency, mild & 0.41$\pm$0.49 & 0.21$\pm$0.41 & -- & 0.00 \\
Aortic insufficiency, moderate & 0.09$\pm$0.29 & 0.07$\pm$0.26 & -- & 0.00 \\
Aortic insufficiency, severe & 0.01$\pm$0.09 & 0.00$\pm$0.00 & 0.939 & 0.00 \\
Aortic insufficiency, trace/trivial & 0.26$\pm$0.44 & 0.13$\pm$0.34 & -- & 0.00 \\
Aortic valve mean gradient (echo, mm Hg) & 40.19$\pm$15.00 & 43.96$\pm$12.36 & 0.284 & 83.33 \\
Aortic valve VTI (cm) & 96.34$\pm$23.65 & 102.54$\pm$20.51 & 0.274 & 83.33 \\
\bottomrule
\end{tabular}
}
\end{table}

\subsection{St. Vincent's Hospital} \label{app_subsec: st_vinc}

\begin{table}[H]
\centering
\footnotesize
\caption{Baseline characteristics of the St.~Vincent’s Hospital cohort. Values are mean~$\pm$~SD, with overall missingness and $p$-values comparing TAVR and SAVR patients.}
\label{tab:baseline_stvincents}
\resizebox{\textwidth}{!}{
\begin{tabular}{lcccc}
\toprule
\textbf{Variable} & \textbf{TAVR} & \textbf{SAVR} & \textbf{$p$-value} & \textbf{Missing (\%)} \\
\midrule
\multicolumn{5}{l}{\textbf{Demographics}} \\
Sex (female = 1) & 0.32$\pm$0.47 & 0.22$\pm$0.42 & 0.20 & 0.0 \\
Age (years) & 76.99$\pm$7.86 & 62.58$\pm$10.55 & $<10^{-9}$ & 0.0 \\
Body Mass Index (kg/m$^2$) & 33.20$\pm$32.48 & 29.50$\pm$5.16 & 0.09 & 0.0 \\
STS Predicted Risk of Mortality (\%) & 2.15$\pm$0.81 & 1.13$\pm$0.72 & $<10^{-9}$ & 0.0 \\
\addlinespace
\multicolumn{5}{l}{\textbf{Prior Medical Conditions}} \\
Porcelain aorta & 0.01$\pm$0.12 & 0.00$\pm$0.00 & 0.05 & 0.0 \\
Peripheral arterial disease (history) & 0.09$\pm$0.29 & 0.00$\pm$0.00 & $<10^{-6}$ & 0.0 \\
Transient ischaemic attack (history) & 0.04$\pm$0.20 & 0.03$\pm$0.17 & 0.67 & 0.0 \\
Prior myocardial infarction & 0.13$\pm$0.34 & 0.14$\pm$0.35 & 0.94 & 0.0 \\
Prior percutaneous coronary intervention & 0.27$\pm$0.44 & 0.17$\pm$0.38 & 0.15 & 0.0 \\
Home supplemental oxygen therapy & 0.01$\pm$0.11 & 0.00$\pm$0.00 & 0.08 & 0.0 \\
Current dialysis & 0.01$\pm$0.09 & 0.00$\pm$0.00 & 0.16 & 0.0 \\
Hypertension (history) & 0.82$\pm$0.38 & 0.83$\pm$0.38 & 0.86 & 0.0 \\
Diabetes mellitus & 0.31$\pm$0.46 & 0.25$\pm$0.44 & 0.43 & 0.0 \\
Cerebrovascular disease (history) & 0.14$\pm$0.35 & 0.11$\pm$0.32 & 0.60 & 0.0 \\
Coronary artery disease (history) & 0.14$\pm$0.35 & 0.03$\pm$0.17 & 0.002 & 0.0 \\
Chronic obstructive pulmonary disease (history) & 0.16$\pm$0.37 & 0.17$\pm$0.38 & 0.92 & 0.0 \\
Immunosuppression (history) & 0.05$\pm$0.22 & 0.08$\pm$0.28 & 0.52 & 0.0 \\
Recent heart failure (within 2~weeks) & 0.70$\pm$0.46 & 0.00$\pm$0.00 & $<10^{-70}$ & 0.0 \\
Left main coronary disease & 0.01$\pm$0.09 & 0.00$\pm$0.00 & 0.16 & 0.0 \\
Proximal LAD disease & 0.08$\pm$0.27 & 0.00$\pm$0.00 & $<10^{-6}$ & 0.0 \\
Intra-aortic balloon pump support & 0.00$\pm$0.00 & 0.00$\pm$0.00 & --- & 0.0 \\
Serum albumin (g/dL) & 4.28$\pm$0.32 & 4.20$\pm$0.51 & 0.37 & 18.7 \\
\addlinespace
\multicolumn{5}{l}{\textbf{CTA Measurements}} \\
Right coronary ostium height (mm) & --- & --- & --- & 100.0 \\
Left coronary ostium height (mm) & --- & --- & --- & 100.0 \\
Aortic root angulation (°) & --- & --- & --- & 100.0 \\
Sinotubular junction diameter (mm) & --- & --- & --- & 100.0 \\
Sinus of Valsalva diameter (mm) & --- & --- & --- & 100.0 \\
Aortic annulus major diameter (mm) & --- & --- & --- & 100.0 \\
Aortic annulus minor diameter (mm) & --- & --- & --- & 100.0 \\
\addlinespace
\multicolumn{5}{l}{\textbf{Echocardiography Measurements}} \\
Left ventricular ejection fraction (\%) & 59.90$\pm$8.13 & --- & --- & 12.5 \\
Aortic valve mean gradient (mmHg, highest documented) & 43.58$\pm$13.43 & 43.64$\pm$10.07 & 0.98 & 3.9 \\
Aortic valve peak velocity (m/s) & 3.89$\pm$0.70 & 4.24$\pm$0.63 & 0.008 & 7.9 \\
Aortic insufficiency, mild & 0.34$\pm$0.48 & 0.25$\pm$0.44 & --- & 0.0 \\
Aortic insufficiency, moderate & 0.16$\pm$0.36 & 0.17$\pm$0.38 & --- & 0.0 \\
Aortic insufficiency, severe & 0.01$\pm$0.11 & 0.22$\pm$0.42 & --- & 0.0 \\
Aortic insufficiency, trace/trivial & 0.22$\pm$0.42 & 0.22$\pm$0.42 & --- & 0.0 \\
Aortic valve mean gradient (echo, mmHg) & 43.25$\pm$10.13 & 37.61$\pm$9.52 & 0.07 & 80.7 \\
Aortic valve VTI (cm) & 97.26$\pm$13.15 & 85.74$\pm$17.30 & 0.03 & 80.7 \\
\bottomrule
\end{tabular}
}
\vspace{-0.5em}
\end{table}

\newpage
\section{Effect of Prognostic Matching} \label{app_sec: prog_matching}

\begin{figure}[H]
\centering
\includegraphics[width=0.8\textwidth]{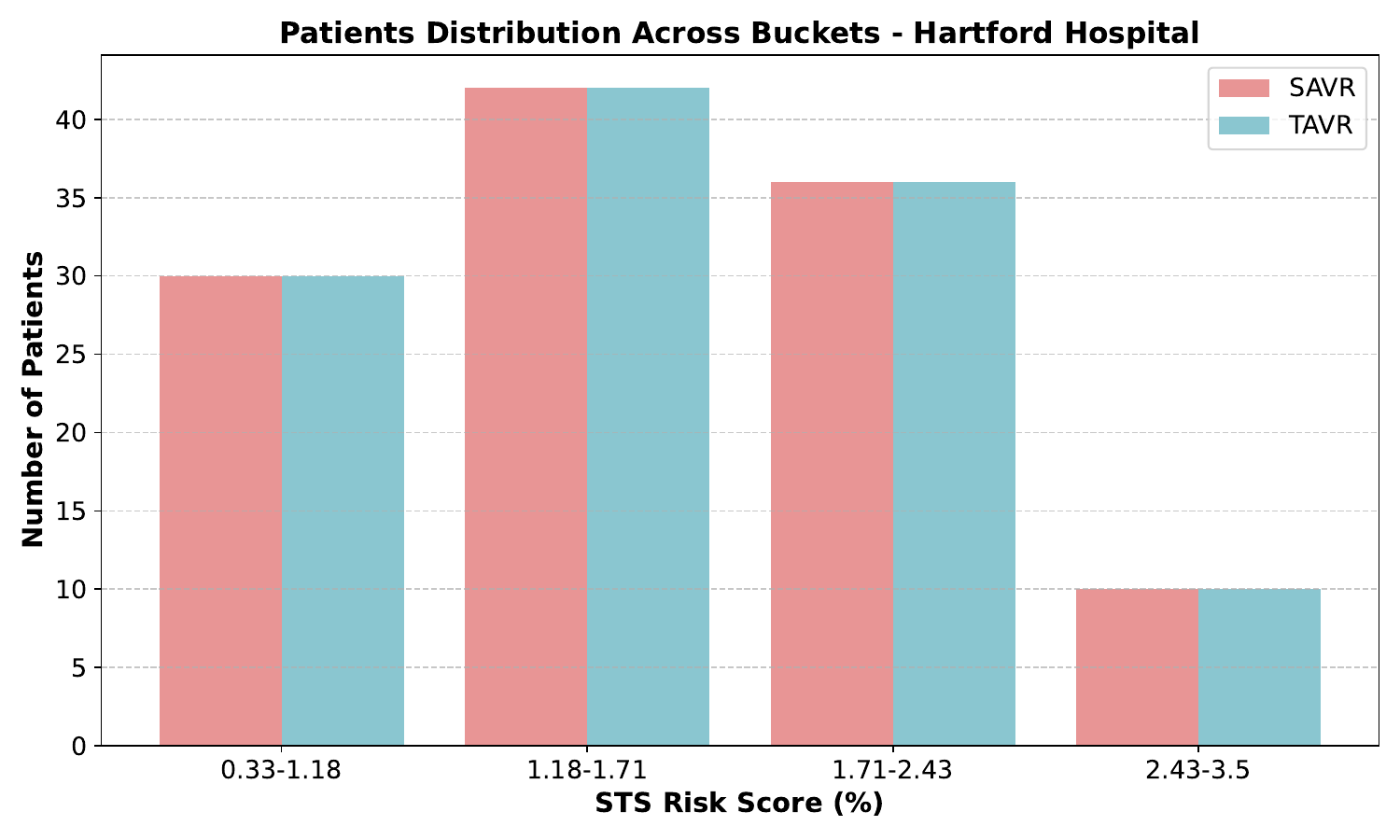}
\caption{Distribution of patients across Society of Thoracic Surgeons (STS) risk score strata in the Hartford (training) cohort after prognostic matching. Bucket boundaries were selected to achieve approximately equal numbers of patients per subgroup, ensuring stable estimates. These cutoff values therefore do not correspond to rounded clinical thresholds.}
\label{fig:dist_post_matching}
\vspace{-0.5em}
\end{figure}

To assess covariate balance between treatment groups, we computed the absolute standardized mean difference (SMD) for each variable. For continuous features, the SMD is defined as:
\[
\text{SMD} = 
\frac{|\mu_{\text{TAVR}} - \mu_{\text{SAVR}}|}
{\sqrt{\frac{1}{2}\left(\sigma_{\text{TAVR}}^2 + \sigma_{\text{SAVR}}^2\right)}}.
\]
For binary variables, proportions replace means. Absolute SMDs are not influenced by sample size and provide a consistent measure of covariate balance, with values below 0.1 generally considered acceptable.

\begin{figure}[H]
\centering
\includegraphics[width=0.48\textwidth]{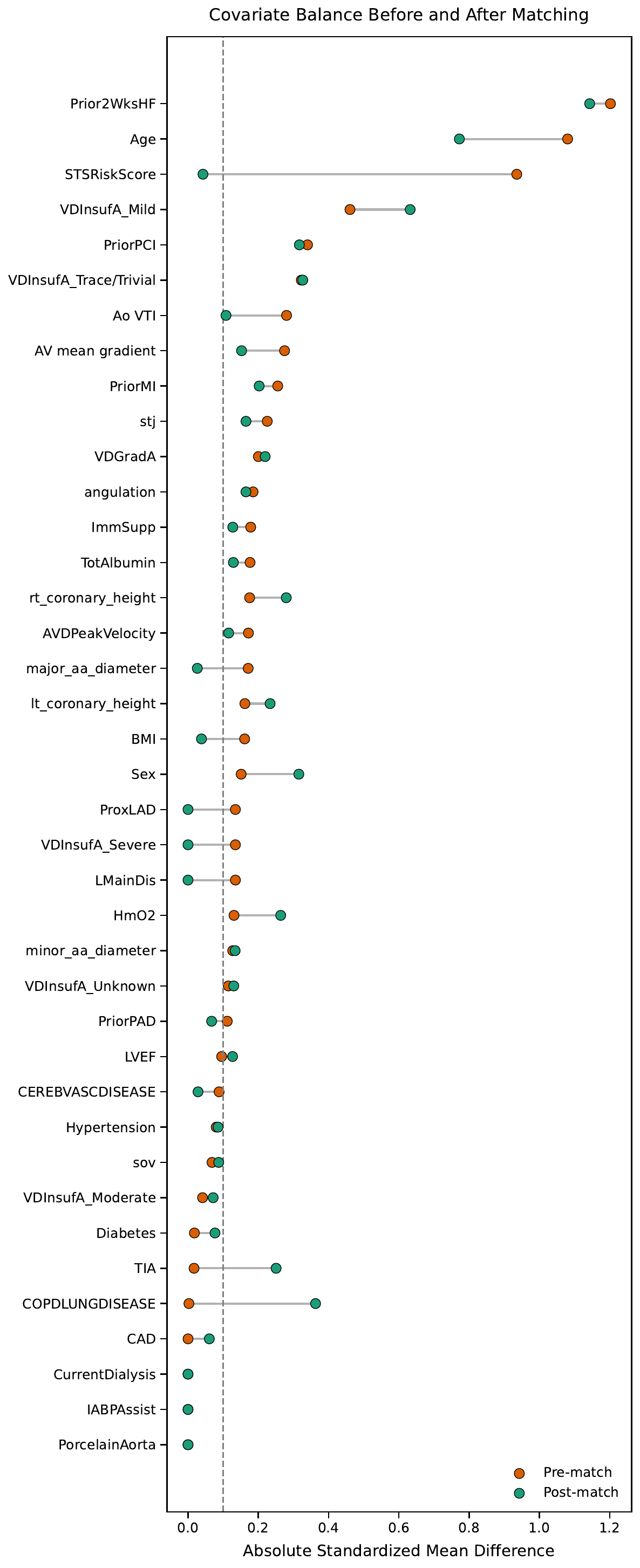}
\caption{
Absolute standardized mean differences (SMDs) for covariates before and after prognostic matching in the Hartford Hospital cohort.
Each line represents one covariate, connecting its pre- and post-match imbalance.
The dashed vertical line marks the conventional SMD threshold of 0.1, below which covariate imbalance is considered negligible.
}
\label{fig:love_plot_full}
\vspace{-0.5em}
\end{figure}

\begin{table}[H]
\centering
\caption{Baseline characteristics of the Hartford Hospital cohort after 1:1 matching. Values are mean~$\pm$~SD, with overall missingness and $p$-values comparing TAVR and SAVR patients.}
\label{tab:baseline_cohorts_post_matching}
\resizebox{\textwidth}{!}{
\begin{tabular}{lcccc}
\toprule
\textbf{Variable} & \textbf{TAVR (mean$\pm$SD)} & \textbf{SAVR (mean$\pm$SD)} & \textbf{$p$-value} & \textbf{Missing (\%)} \\
\midrule
\multicolumn{5}{l}{\textbf{Demographics}}\\
Sex (female = 1) & 0.31$\pm$0.47 & 0.47$\pm$0.50 & 0.016 & 0.00 \\
Age (years) & 83.10$\pm$5.91 & 78.26$\pm$6.61 & $<10^{-8}$ & 0.00 \\
Body Mass Index (kg/m$^2$) & 27.61$\pm$4.80 & 27.84$\pm$7.06 & 0.769 & 0.00 \\
STS Predicted Risk of Mortality (\%) & 1.64$\pm$0.59 & 1.61$\pm$0.62 & 0.744 & 0.00 \\
\addlinespace
\multicolumn{5}{l}{\textbf{Prior Medical Conditions}}\\
Porcelain aorta & 0.00$\pm$0.00 & 0.00$\pm$0.00 & -- & 0.00 \\
Peripheral arterial disease (history) & 0.08$\pm$0.27 & 0.06$\pm$0.24 & 0.606 & 0.00 \\
Transient ischaemic attack (history) & 0.01$\pm$0.09 & 0.05$\pm$0.22 & 0.056 & 0.00 \\
Prior myocardial infarction & 0.09$\pm$0.29 & 0.04$\pm$0.20 & 0.121 & 0.00 \\
Prior percutaneous coronary intervention & 0.14$\pm$0.35 & 0.05$\pm$0.22 & 0.016 & 0.00 \\
Home supplemental oxygen therapy & 0.00$\pm$0.00 & 0.03$\pm$0.18 & 0.045 & 0.00 \\
Current dialysis & 0.00$\pm$0.00 & 0.00$\pm$0.00 & -- & 0.00 \\
Hypertension (history) & 0.89$\pm$0.31 & 0.92$\pm$0.28 & 0.512 & 0.00 \\
Diabetes mellitus & 0.28$\pm$0.45 & 0.25$\pm$0.43 & 0.556 & 0.00 \\
Cerebrovascular disease (history) & 0.10$\pm$0.30 & 0.09$\pm$0.29 & 0.827 & 0.00 \\
Coronary artery disease (history) & 0.09$\pm$0.29 & 0.08$\pm$0.27 & 0.642 & 0.00 \\
Chronic obstructive pulmonary disease (history) & 0.16$\pm$0.37 & 0.31$\pm$0.47 & 0.006 & 0.00 \\
Immunosuppression (history) & 0.09$\pm$0.29 & 0.06$\pm$0.24 & 0.329 & 0.00 \\
Recent heart failure (within 2 weeks) & 0.42$\pm$0.49 & 0.01$\pm$0.09 & $<10^{-14}$ & 0.00 \\
Left main coronary disease & 0.00$\pm$0.00 & 0.00$\pm$0.00 & -- & 0.00 \\
Proximal LAD disease & 0.00$\pm$0.00 & 0.00$\pm$0.00 & -- & 0.00 \\
Intra-aortic balloon pump support & 0.00$\pm$0.00 & 0.00$\pm$0.00 & -- & 0.00 \\
Serum albumin (g/dL) & 3.83$\pm$0.45 & 3.77$\pm$0.46 & 0.323 & 0.00 \\
\addlinespace
\multicolumn{5}{l}{\textbf{CTA Measurements}}\\
Right coronary ostium height (mm) & 18.86$\pm$4.07 & 17.73$\pm$4.05 & 0.033 & 0.00 \\
Left coronary ostium height (mm) & 17.39$\pm$4.17 & 16.42$\pm$4.12 & 0.074 & 0.00 \\
Aortic root angulation (°) & 48.89$\pm$7.03 & 50.15$\pm$8.25 & 0.207 & 0.00 \\
Sinotubular junction diameter (mm) & 28.81$\pm$2.60 & 29.24$\pm$2.60 & 0.206 & 0.00 \\
Sinus of Valsalva diameter (mm) & 35.12$\pm$3.99 & 34.78$\pm$3.80 & 0.503 & 0.00 \\
Aortic annulus major diameter (mm) & 27.65$\pm$3.09 & 27.56$\pm$3.41 & 0.839 & 0.00 \\
Aortic annulus minor diameter (mm) & 22.80$\pm$3.16 & 22.35$\pm$3.42 & 0.304 & 0.00 \\
\addlinespace
\multicolumn{5}{l}{\textbf{Echocardiography Measurements}}\\
Left ventricular ejection fraction (\%) & 63.15$\pm$9.31 & 61.91$\pm$10.36 & 0.331 & 0.00 \\
Aortic valve mean gradient (mm Hg) & 30.99$\pm$6.57 & 31.98$\pm$6.40 & 0.243 & 0.00 \\
Aortic valve peak velocity (m/s) & 4.41$\pm$1.01 & 4.52$\pm$0.97 & 0.375 & 0.00 \\
Aortic insufficiency, mild & 0.45$\pm$0.50 & 0.17$\pm$0.38 & $<10^{-5}$ & 0.00 \\
Aortic insufficiency, moderate & 0.07$\pm$0.25 & 0.05$\pm$0.22 & 0.583 & 0.00 \\
Aortic insufficiency, severe & 0.00$\pm$0.00 & 0.00$\pm$0.00 & -- & 0.00 \\
Aortic insufficiency, trace/trivial & 0.25$\pm$0.44 & 0.13$\pm$0.33 & 0.013 & 0.00 \\
Aortic valve velocity time integral (cm) & 84.89$\pm$12.79 & 86.26$\pm$12.49 & 0.407 & 0.00 \\
\bottomrule
\end{tabular}
}
\end{table}

\begin{figure}[t]
\centering
\includegraphics[width=0.8\textwidth]{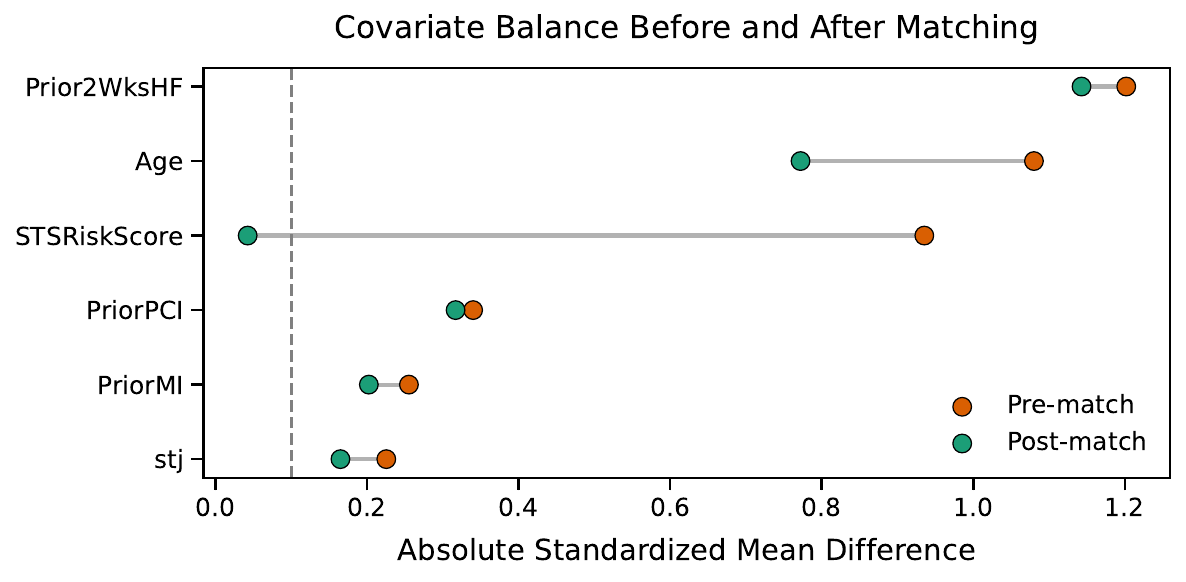}
\caption{
Absolute standardized mean differences (SMDs) for the set of previously statistically different covariates before and after prognostic matching in the Hartford Hospital cohort.
Each line represents one covariate, connecting its pre- and post-match imbalance.
The dashed vertical line marks the conventional SMD threshold of 0·1, below which covariate imbalance is considered negligible.
}
\label{fig:love_plot_rest}
\vspace{-0.5em}
\end{figure}

\newpage
\section{Effect of Sample Weighting} \label{app_sec: sample_weight}

\begin{figure}[H]
\centering
\includegraphics[width=0.8\textwidth]{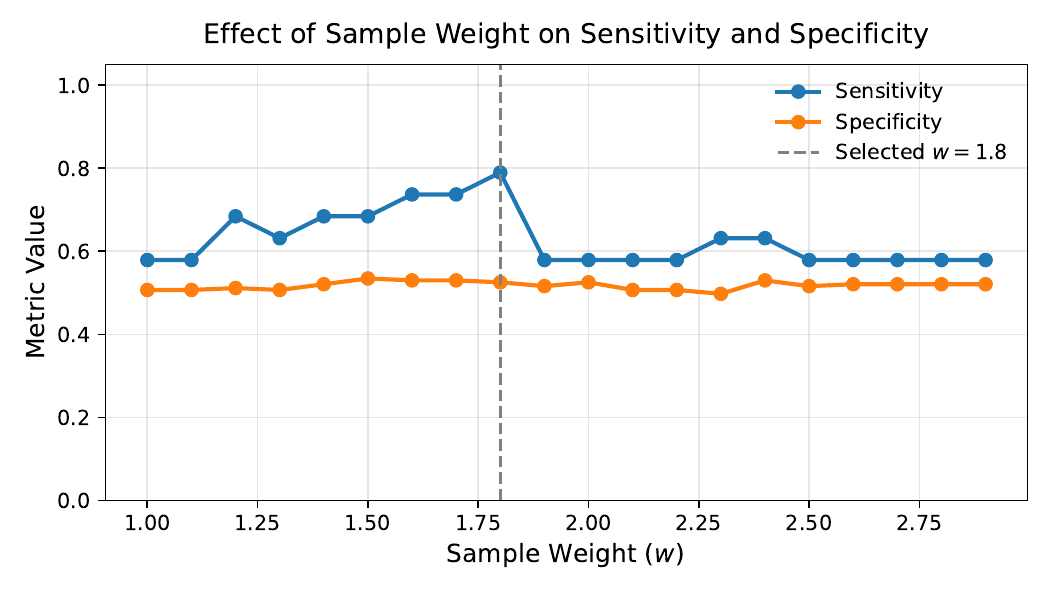}
\caption{
Sensitivity and Specificity across different values of sample weight $w$.
}
\label{fig:sample_weighting}
\vspace{-0.5em}
\end{figure}

Figure \ref{fig:sample_weighting} shows the effect of increasing the sample weight that upweights SAVR patients with a negative outcome, and TAVR patients with a positive outcome. The goal of this process is to improve balance across cohorts. The weight selected is $w=1.8$, since it resulted in the tree with the highest average sensitivity and specificity, that is also medically sound. We observe that as we keep increasing $w$ above 1.8, the sensitivity of the trees decreases.

\end{appendices}